\documentclass[11pt]{article}
\usepackage{float}
\usepackage[preprint]{acl}
\usepackage{times}
\usepackage{latexsym}
\usepackage[T1]{fontenc}
\usepackage[utf8]{inputenc}
\usepackage{microtype}
\usepackage{inconsolata}
\usepackage{graphicx}
\usepackage{booktabs}
\usepackage{amsmath}
\usepackage{amssymb}
\usepackage{enumitem}
\usepackage{algorithm}
\usepackage{algpseudocode}

\title{PatchBoard: Schema-Grounded State Mutation for Reliable and Auditable LLM Multi-Agent Collaboration}

\author{%
  \textbf{Shuyu Zhang \quad Yaqi Shi \quad Jiarui Zhang \quad Yanxiao Zhao \quad Lu Wang\thanks{Corresponding author.} }\\
  School of Computer Science and Technology \\
  Xidian University \\
  Xi'an, China \\
  \texttt{wanglu@xidian.edu.cn}
}

\newcommand{\method}{PatchBoard}
\newcommand{\state}{\mathcal{S}}
\newcommand{\schema}{\Sigma}
\newcommand{\metaschema}{\Sigma_{\mathrm{meta}}}
\newcommand{\rules}{\mathcal{R}}
\newcommand{\patch}{\Delta}
\newcommand{\view}{\mathcal{V}}

\begin{document}
\maketitle

\begin{abstract}
LLM multi-agent systems often coordinate through natural-language dialogue or loosely structured shared memory, making intermediate state difficult to validate, attribute, and audit. We introduce PatchBoard, a schema-grounded collaboration architecture that replaces inter-agent dialogue with validated JSON Patch mutations over a shared structured state. An Architect agent constructs a task-specific schema and workflow rules, while a deterministic kernel validates each proposed state mutation against schema constraints, role-specific write contracts, and runtime invariants before committing it transactionally. On 630 matched ALFWorld episodes, PatchBoard achieves an 84.6\% success rate, compared with 30.8\% for LangGraph and 61.6\% for Flock, while reducing tokens per successful task to 45.5k, compared with 368.3k and 64.2k, respectively.
\end{abstract}

\section{Introduction}

Large language models are increasingly used as autonomous agents that plan, reason, call tools, interact with environments, and revise their behavior through feedback \citep{yao2023react,schick2023toolformer,shinn2023reflexion,yao2023tree,wang2023voyager}. As tasks become longer and more compositional, a natural extension is to organize multiple agents into role-specialized teams. In such systems, different agents coordinate through multi-turn interaction. Representative systems such as AutoGen, CAMEL, ChatDev, MetaGPT, and AgentVerse show that multi-agent collaboration can improve task decomposition and support complex workflows across reasoning, software engineering, simulation, and tool-use settings \citep{wu2024autogen,li2023camel,qian-etal-2024-chatdev,hong2024metagpt,chen2024agentverse,wang2024surveyagent}. The dominant coordination interface in these systems is natural language, which is attractive because it matches the native input-output format of LLMs and makes agent communication flexible and expressive.

However, natural-language communication becomes a fragile substrate for long-horizon and stateful collaboration. Dialogue histories grow with the number of turns, mix task facts with meta-discussion and repair attempts, and often leave unclear which intermediate outputs should be treated as committed state. A downstream agent may read an unverified observation, stale plan, malformed intermediate claim, or failed repair attempt as if it were reliable task state. Once such information enters the shared context, later agents can amplify the error through additional reasoning, tool calls, or environment actions. This problem is especially harmful in collaborative settings because failure is not localized to one model call. A polluted intermediate state can silently affect all subsequent agents.

Existing work attempts to address these issues by making agent coordination more explicit. Some systems use workflow graphs, planners, or verification functions to constrain execution \citep{langgraph2024,zhang2024aflow}. Others ask agents to generate or reuse executable programs and skills, enabling compact and compositional control policies \citep{wang2023voyager,yang2025codeagents}. Blackboard-style systems instead coordinate agents through shared memory, allowing independent workers to read and update a common state \citep{hayesroth1985blackboard,salemi2025blackboard}. Structured generation methods such as LMQL and Outlines further help models produce outputs that follow specified formats \citep{beurer2023prompting,willard2023outlines}. These approaches improve over unconstrained dialogue, but they leave an important gap. Workflow and code-based methods still require the runtime to trust generated procedures or control logic, while blackboard memory does not by itself determine whether an update is well typed, authorized, non-stale, or safe to commit. Structured output helps with formatting, but formatting alone does not define a system-level boundary between a model suggestion and committed shared state.

This paper proposes \method{}, a schema-grounded communication substrate for reliable and auditable LLM multi-agent collaboration. \method{} replaces open-ended inter-agent dialogue with validated JSON Patch mutations over a shared JSON state \citep{bryan2013rfc}. An Architect agent defines the task schema, worker contracts, context budgets, and workflow rules \citep{bourhis2017json}, while a deterministic kernel validates and transactionally commits only authorized state updates. This makes collaboration explicit, attributable, and replayable, preventing malformed or unauthorized outputs from silently entering shared memory. On ALFWorld \citep{shridhar2021alfworld}, \method{} achieves an 84.6\% success rate over 630 matched episodes, compared with 30.8\% for LangGraph and 61.6\% for Flock. It also achieves the lowest normalized cost, requiring 45.5k tokens per successful task compared with 368.3k for LangGraph and 64.2k for Flock.

We make the following contributions:
\begin{itemize}[leftmargin=*]
    \item We formulate LLM multi-agent collaboration as validated mutation over a shared structured state, using a restricted JSON Patch interface to make inter-agent communication explicit, typed, and auditable.

    \item We design a deterministic kernel that validates proposed updates, enforces schema and role-specific write constraints, constructs budgeted context views, commits accepted patches transactionally, and records replayable transaction logs.

    \item We build a full \method{} prototype and evaluate it on long-horizon interaction tasks, with blackboard controls, ablations, sensitivity analyses, fault injection, and a diagnostic QA study that clarifies the boundary between structural validation and semantic support.
\end{itemize}


\section{Related Work}

\paragraph{LLM multi-agent coordination.}
Recent LLM multi-agent systems decompose complex tasks into role-specialized agents that communicate, critique, and coordinate through explicit interaction protocols. AutoGen and AgentScope provide general-purpose infrastructures for composing conversational or message-passing agents \citep{wu2024autogen,agentscope}, while ChatDev and MetaGPT instantiate role-based collaboration for software development through chat chains or SOP-style workflows \citep{qian-etal-2024-chatdev,hong2024metagpt}. Recent analyses further show that coordination failures, ambiguous handoffs, and redundant communication remain central challenges in LLM-based multi-agent systems \citep{cemri2025why}. This line of work motivates treating the communication substrate itself as a first-class object in multi-agent design.

\paragraph{Structured workflows and verification-aware orchestration.}
A growing body of work makes agent execution more explicit through graphs, state machines, language-model programs, or automatically optimized workflows. LangGraph represents agent applications as stateful graphs \citep{langgraph2024}; StateFlow formulates LLM task solving as state-driven workflows \citep{wu2024stateflow}; DSPy abstracts LM pipelines as declarative programs that can be optimized \citep{khattab2024dspy}; and SGLang targets efficient execution of structured language-model programs \citep{zheng2024sglang}. Recent systems such as AFlow and VeriMAP further explore automated workflow generation and verification-aware multi-agent planning \citep{zhang2024aflow,xu2026verification}. These works show the value of moving agent coordination from free-form chat toward explicit control structures.

\paragraph{Shared memory, blackboards, and agent memory.}
Blackboard architectures provide a classical mechanism for coordinating independent knowledge sources through a shared state \citep{hayesroth1985blackboard,penny1986blackboard}. This idea has recently reappeared in LLM multi-agent systems, where blackboard-style memory supports dynamic agent selection, shared information discovery, and event-driven collaboration \citep{han2025blackboard,salemi2025blackboard}. In parallel, agent memory systems study how long-term observations can be stored, linked, and retrieved to support persistent behavior across interactions \citep{packer2023memgpt,xu2025amem}. These works highlight the importance of persistent shared state, while leaving open how such state should be updated, authorized, and audited during long-horizon collaboration.

\paragraph{Structured outputs, transactions, and semantic verification.}
Structured generation techniques constrain LLM outputs with grammars, schemas, or programming interfaces, reducing format errors in machine-consumed outputs \citep{beurer2023prompting,willard2023outlines,zheng2024sglang,geng2025jsonschemabench}. Related systems work brings stronger runtime guarantees into LLM agents: SagaLLM studies context management, validation, and transaction guarantees for multi-agent planning \citep{chang2025sagallm}, while recent runtime-governance work emphasizes path-dependent policy enforcement for autonomous agents \citep{kaptein2026runtime}. Finally, evidence-grounded QA and fact-checking benchmarks such as HotpotQA, FEVER, and MuSiQue evaluate whether generated claims are supported by evidence \citep{yang2018hotpotqa,thorne2018fever,trivedi2022musique}. Together, these lines connect structured output control, transactional execution, and semantic verification.

Overall, \method{} advances this direction by making collaboration a sequence of schema-grounded, role-authorized, replayable state mutations, giving multi-agent systems a tighter runtime boundary for reliable and auditable coordination.

\section{Methodology}

\begin{figure*}[t]
\centering
\includegraphics[width=\linewidth]{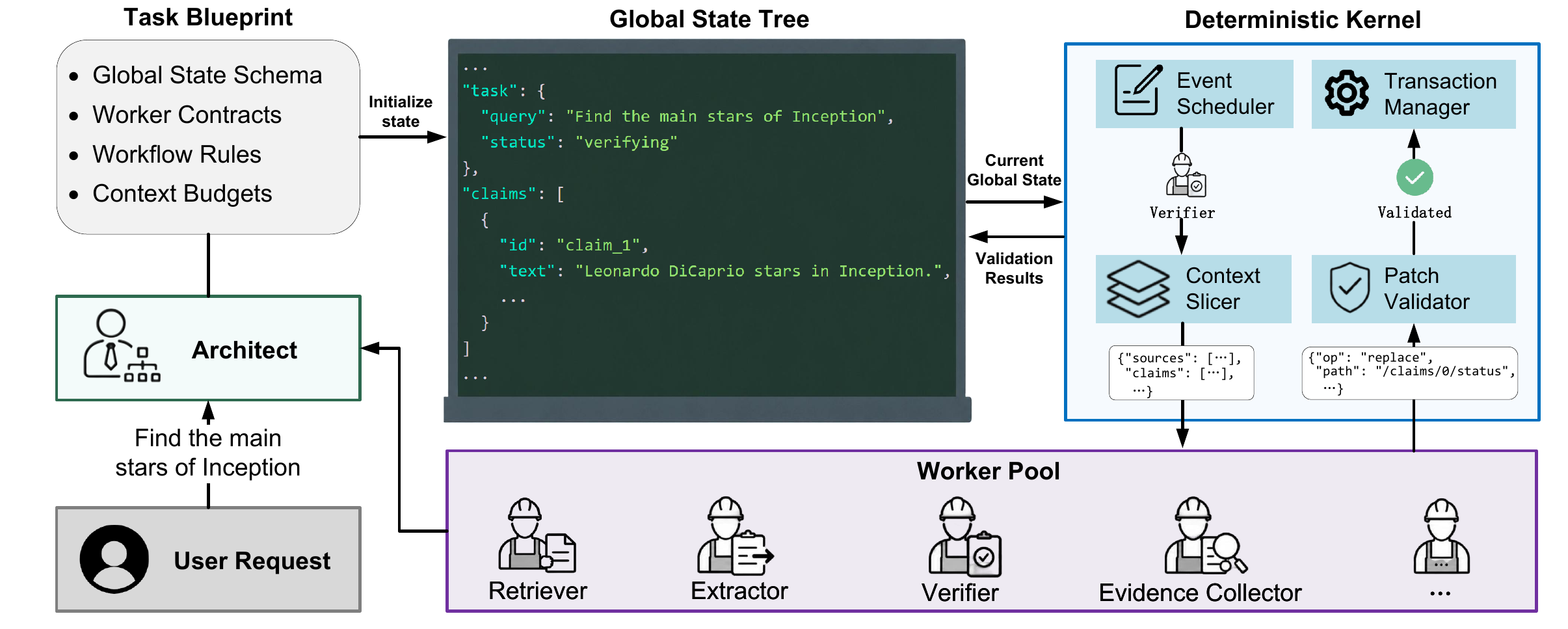}
\caption{\method{} architecture. The Architect compiles a user request into a task blueprint containing the global state schema, worker contracts, workflow rules, and context budgets. The deterministic kernel maintains the global state tree, constructs bounded state views, validates worker-proposed JSON Patches, commits accepted updates transactionally, and schedules subsequent worker invocations. Workers interact through schema-validated mutations to the shared state.}
\label{fig:architecture}
\end{figure*}

\subsection{Method Overview}
\label{sec:method-overview}

\method{} formulates multi-agent collaboration as a closed state-transition loop over a shared structured state. As shown in Figure~\ref{fig:architecture}, an Architect first converts the user request into a task blueprint specifying the global state schema, worker contracts, context budgets, and workflow rules. After the blueprint is validated, runtime coordination is handled by a deterministic kernel, which initializes the global state, constructs bounded worker views, validates JSON Patch proposals, commits accepted patches, records transaction logs, and schedules future workers from committed state events.

Let $\state_t$ denote the global state tree at step $t$. A hand-crafted blueprint meta-schema $\metaschema$ defines the legal structure of Architect-produced blueprints. An accepted blueprint $\mathcal{B}$ instantiates a task-specific schema $\schema$, a set of workers $\mathcal{A}$, and workflow rules $\rules$. The schema $\schema$ defines the valid structure and invariants of $\state_t$, while $\rules$ maps committed state events to future worker invocations. For a worker $a \in \mathcal{A}$, the kernel materializes a bounded view $\view_t^a$ and receives a candidate patch $\patch_t^a$.

The kernel is the only component that can mutate the committed state. Let $W_a$ denote the write contract of worker $a$, and let $\bot$ denote failed patch application. The kernel first applies the candidate patch to a temporary copy of the current state,
\begin{equation}
\label{eq:accept}
\begin{aligned}
\hat{\state}_{t+1}
&= \mathsf{Apply}(\state_t,\patch_t^a),\\
\mathsf{Accept}_t^a
&= \mathsf{Syntax}(\patch_t^a)
   \land \mathsf{Auth}(\patch_t^a,W_a)\\
&\quad \land\; (\hat{\state}_{t+1} \neq \bot)
   \land \mathsf{Valid}_{\schema}(\hat{\state}_{t+1})\\
&\quad \land\; \mathsf{Inv}_{B}
   (\state_t,\patch_t^a,\hat{\state}_{t+1}).
\end{aligned}
\end{equation}
Here $\mathsf{Accept}_t^a$ indicates whether worker $a$'s patch is accepted at step $t$, and $\mathsf{Inv}_{\mathcal{B}}$ denotes the runtime invariants registered by blueprint $\mathcal{B}$. Accepted patches are committed as transactions; rejected patches are logged without changing the committed state. This separates model-generated proposals from accepted system state. The full kernel pseudocode is provided in Appendix~\ref{app:pseudocode}.

\subsection{Architect and Task Blueprint}
\label{sec:architect}

The Architect is invoked once at task initialization. Given a user request, it produces a blueprint $\mathcal{B}$ that defines the collaboration structure before any worker is called. The blueprint contains the task schema $\schema$, worker specifications, context budgets, and workflow rules $\rules$. These fields determine the layout of the shared state, the roles that may participate in the task, the state regions each role may observe or modify, and the events that trigger future worker invocations.

Before execution, the kernel validates the Architect output against the blueprint meta-schema $\metaschema$. The meta-schema constrains the format of the generated blueprint. Worker names must be declared, read and write paths must refer to valid schema locations, context budgets must be finite, and workflow rules must use the restricted trigger-condition-action format supported by the scheduler. A structurally invalid blueprint is rejected before runtime execution begins.

After validation, the accepted blueprint becomes the runtime contract for the task. Each worker specification contains a role instruction, authorized read paths, authorized write paths, a view budget, and patch-format constraints. For example, an evidence collector may read the task query and unresolved claims, append evidence records under \texttt{/evidence/-}, and have no permission to replace verifier-controlled fields such as \texttt{/claims/*/status}. Workflow rules connect committed state changes to future computation. Adding a source may wake an extractor, while adding an unverified claim may wake a verifier.

This design makes the Architect a setup-time planning component. Its output defines the initial collaboration structure, including the state schema, worker roles, context budgets, and workflow rules. After the blueprint is accepted, runtime coordination is fully mediated by the deterministic kernel through patch validation, transactional commits, and event-based scheduling.

\subsection{Schema-Grounded Patch Interface}
\label{sec:patch-interface}

Workers interact with the shared state through a restricted JSON Patch interface \citep{bryan2013rfc}. At each invocation, a worker receives a bounded view $\view_t^a$ and returns a candidate patch $\patch_t^a$ over the global state tree. Each operation is path-addressed, so the intended edit is explicit at the field level. This makes worker outputs easier to validate, attribute, and replay than free-form messages.

The patch interface is grounded in the accepted schema $\schema$. Each path in a candidate patch must refer to a valid schema location, and each value must satisfy the type and field constraints associated with that location. The allowed operation subset is intentionally small. Workers may add newly produced objects, replace fields assigned to their role, and use \texttt{test} operations to express stale-view preconditions. Destructive operations such as \texttt{remove} are disabled by default and can be enabled only for privileged roles.

Role-specific write contracts further constrain the interface. A worker can only propose edits to paths granted by its blueprint specification. For example, an extractor may append draft claims under \texttt{/claims/-}, while a verifier may replace verification fields under \texttt{/claims/*/status}. This path-level separation prevents one role from silently overwriting intermediate products or decisions assigned to another role.

Patch validation happens before any committed state is modified. The kernel parses the candidate patch, checks operation syntax, verifies path authorization, applies the patch to a temporary copy of $\state_t$, and validates the resulting candidate state against $\schema$ and registered invariants. If all checks pass, the patch is committed as a transaction and produces $\state_{t+1}$. If any check fails, the patch is rejected and logged with its rejection reason.

This interface provides a narrow coordination surface for LLM workers. Workers still generate semantic content, and schema validity alone cannot guarantee factual correctness. However, every accepted update has an explicit writer, path, operation, and post-state validation result. This gives downstream workers and auditors a concrete record of how the shared state evolved.

\subsection{Deterministic Kernel}
\label{sec:kernel}

The deterministic kernel is the runtime controller of \method{}. After a blueprint has been accepted, the kernel maintains the committed global state, the event queue, worker budgets, and the transaction log. For each scheduled worker invocation, it constructs the worker input from the current state, calls the worker, receives a candidate patch, and decides whether the proposed update can become part of the committed trajectory.

The kernel validates every worker output before state mutation. Given $\state_t$, worker $a$, and candidate patch $\patch_t^a$, the kernel checks the allowed JSON Patch operation subset and verifies that every target path is covered by the worker's write contract. It then applies the patch to a temporary copy of $\state_t$ and validates the candidate next state against $\schema$ and any registered invariants. The committed state advances to $\state_{t+1}$ only when all validation checks succeed.

Accepted patches are committed transactionally. A committed transaction records the worker id, triggering event, worker-view hash, accepted patch, and resulting state hash. Rejected patches are also recorded with the failed validation stage and rejection reason. Since rejected patches never modify the committed state, malformed outputs, unauthorized writes, and schema-violating edits remain visible in the log without contaminating the shared state.

The scheduler consumes events emitted by committed transactions and matches them against $\rules$. When a rule condition is satisfied, the corresponding worker invocation is added to the event queue. The runtime trajectory therefore depends on accepted state changes rather than free-form inter-worker messages. The same committed state and transaction log provide a replayable account of how the system reached its final state.

The kernel also monitors simple failure signals during execution. It tracks consecutive invalid patches, repeated no-op edits, exhausted worker budgets, and repeated state hashes. When a configured threshold is reached, the kernel may stop the branch, wake a verifier, reduce the available view, or terminate the task. These policies are deterministic functions of the transaction log and the accepted blueprint.

\subsection{Budgeted Context Views}
\label{sec:context-views}

The global state tree may grow as workers add sources, claims, plans, evidence records, and verification results. Passing the full state to every worker increases context cost and exposes irrelevant fields to roles that do not need them. \method{} therefore makes view construction a kernel responsibility. Before invoking worker $a$, the kernel materializes a bounded view $\view_t^a$ from $\state_t$ according to the worker's read contract and context budget.

A view contains the state fields required by the worker's role, the relevant schema fragment, unresolved dependencies, and recent rejection feedback associated with the same worker or state region. Large collections are represented through compact summaries and stable handles. For example, a verifier may receive the task query, a small set of unresolved claims, their linked evidence handles, and the schema fragment for claim-status updates, while unrelated worker outputs remain outside its view.

The context budget limits the size of the materialized view. When the authorized state region exceeds this budget, the kernel prioritizes active task fields, required schema fields, and recently changed objects. Older or lower-priority collections are compressed into summaries that preserve identifiers and provenance. A worker that needs additional information can propose a typed expansion request through the same patch interface, allowing the kernel to page in specific handles in a later invocation.

This design keeps context selection explicit and auditable. The transaction log records the view hash used for each worker call, so accepted and rejected patches can be traced back to the state slice that produced them. Budgeted views also reduce accidental role leakage, since workers only observe paths allowed by their read contracts. As a result, \method{} controls both what a worker may modify and what information the worker may condition on during patch generation.

\subsection{Structural Properties}
\label{sec:properties}

The preceding components give \method{} several structural properties, conditioned on a valid blueprint and a correct kernel implementation. Since every runtime update passes through the patch validator inside the deterministic kernel, committed states preserve the accepted task schema. If $\state_t$ satisfies $\schema$, the kernel commits a transition only after applying the candidate patch to a temporary copy and validating the resulting state against $\schema$ and registered invariants. Malformed patches, type errors, missing required fields, and schema-violating updates are rejected before they can modify the committed state.

Worker effects are isolated by path-level contracts defined in the blueprint. Each worker can propose patches only to paths authorized by its role, and the kernel checks these paths before applying the patch. This prevents one role from silently overwriting fields assigned to another role. It also makes each accepted update attributable to a specific worker invocation, since the transaction records the worker id, the viewed state slice, the proposed patch, and the validation outcome.

Accepted trajectories are replayable at the patch level. The transaction manager records accepted and rejected patches, input view hashes, rejection reasons, and resulting state hashes. Given the initial state, the accepted blueprint, and the committed transaction log, the sequence of accepted state transitions can be reconstructed without resampling worker outputs. This supports debugging and auditability, while the event scheduler makes downstream worker invocations deterministic with respect to committed state events and workflow rules.

\section{Experimental Setup}
\label{sec:experimental-setup}

\subsection{Evaluation Goals}

The experiments evaluate whether schema-grounded state mutation improves task success and normalized cost, whether the gains can be explained by shared memory alone, and which \method{} components contribute most to the observed behavior. We also include controlled fault injection and a diagnostic QA setting to distinguish structural state validity from semantic support.

\subsection{Benchmarks and Systems}

The primary benchmark is ALFWorld \citep{shridhar2021alfworld}. We use 126 matched gamefiles stratified across six ALFWorld task types and run each gamefile with 5 independent execution seeds, yielding 630 episodes per system. All systems use the same gamefiles, seeds, model, decoding configuration, environment step budget, and timeout. An episode is counted as successful only when the environment reaches the target terminal condition within 20 environment steps.

The main ALFWorld comparison includes \method{}, LangGraph \citep{langgraph2024}, and Flock \citep{flockdocs2025}. LangGraph serves as a graph-based workflow baseline with explicit nodes and shared state, while Flock serves as a blackboard-based multi-agent baseline following its public repository and documentation. All systems use the same ALFWorld observation-action interface, model, decoding setting, step budget, and timeout. Detailed baseline configurations are provided in Appendix~\ref{app:baseline_setting}.

The secondary benchmark is a HotpotQA diagnostic \citep{yang2018hotpotqa} with 240 matched prepared validation examples. This setting evaluates evidence-grounded claim propagation rather than long-horizon environment interaction. It is used to diagnose whether structurally valid workflows reduce unsupported factual claims. Full HotpotQA results are reported in Appendix~\ref{app:hotpotqa}.

Unless otherwise stated, all experiments use Qwen-plus~\citep{yang2025qwen3technicalreport} with temperature 0 and a 60-second timeout. Full configuration details are summarized in Appendix~\ref{app:exp-params}.

\subsection{Metrics and Estimation}

For ALFWorld, we measure task success, environment steps, total token usage, and tokens per success. Tokens per success is computed as mean total tokens divided by success rate. Token totals include prompt and completion tokens from all LLM calls, covering blueprint generation, worker calls, schema/context views, and patch-format instructions for \method{}, as well as the corresponding planner, controller, worker, setup, and repair calls for the baselines. Appendix~\ref{app:cost} gives a concrete \method{} running example and a component-level token cost breakdown.

For HotpotQA, we use answer accuracy, unsupported claim rate, evidence coverage, and verified claim precision as diagnostic metrics. Unsupported claim rate measures schema-valid claims that lack sufficient evidence, while invalid state rate measures malformed or unauthorized intermediate updates that enter committed state. Where applicable, proportion metrics use Wilson intervals and continuous metrics use paired bootstrap intervals over matched examples.
\begin{figure*}[!t]
\centering
\includegraphics[width=0.8\linewidth]{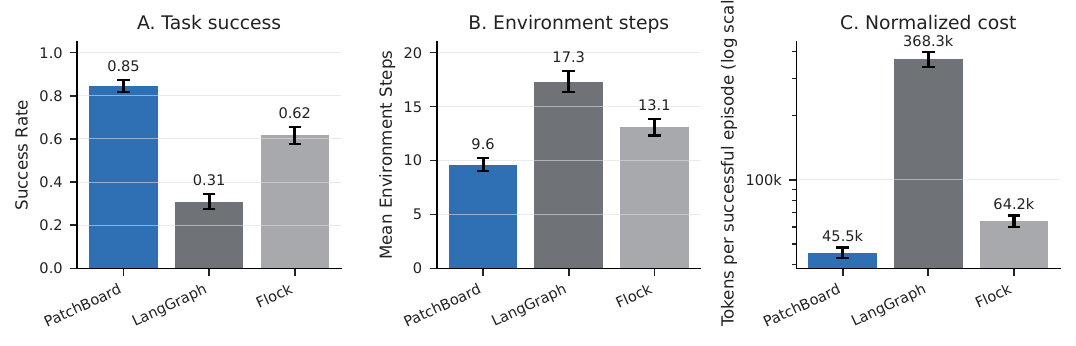}
\caption{Main ALFWorld comparison under matched gamefiles and execution seeds. Tokens per successful episode use a log scale.}
\label{fig:alfworld-main}
\end{figure*}
We report 95\% confidence intervals for all main metrics. Proportion metrics use Wilson intervals, including success rate, answer accuracy, unsupported claim rate, invalid state rate, evidence coverage, verified claim precision, fault contamination rates, and cycle halt rate. Continuous metrics use paired bootstrap intervals over matched task and seed identifiers, including mean steps, mean solved steps, mean total tokens, tokens per success, and tokens per answer.

\subsection{Fault Injection}

We inject 200 instances of each fault type into each system. The injected faults cover Invalid JSON, Bad Path/Type, Unauthorized Write, False Claim, and Cycle Halt. The first three faults test whether malformed or unauthorized updates contaminate committed state. False Claim tests whether schema-valid but unsupported content is accepted. Cycle Halt tests whether repeated no-op or oscillatory trajectories are stopped.

\section{Results and Analysis}
\label{sec:results}

\subsection{Main ALFWorld Results}
\label{sec:main-alfworld}

Figure~\ref{fig:alfworld-main} summarizes the main ALFWorld comparison in terms of task success, environment steps, and tokens per success.

\method{} solves 533/630 episodes, compared with 194/630 for LangGraph and 388/630 for Flock. It also achieves the lowest tokens per success, indicating that validated state mutation improves both task success and normalized cost in this matched ALFWorld setting.

\subsection{Blackboard Controls}
\label{sec:blackboard-controls}

\begin{figure}[H]
\centering
\includegraphics[width=\linewidth]{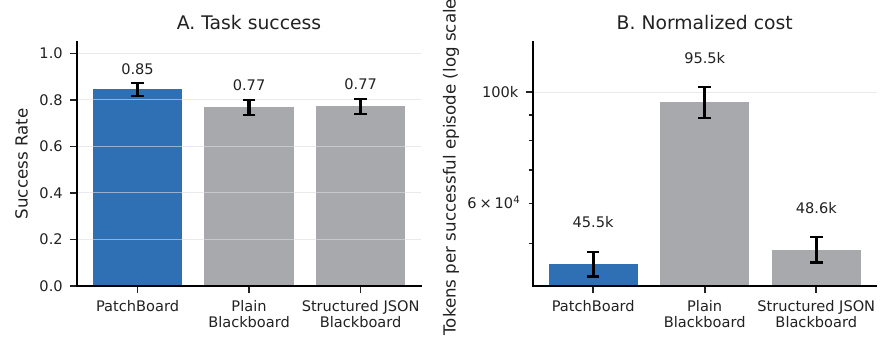}
\caption{Blackboard controls under matched ALFWorld episodes.}
\label{fig:blackboard-controls}
\end{figure}

Blackboard controls test whether the improvement comes merely from adding shared memory. Figure~\ref{fig:blackboard-controls} compares \method{} with plain and structured blackboards under the same matched ALFWorld setting.

Both blackboard controls solve fewer episodes than \method{}. The plain blackboard also incurs a much higher cost per successful task, while the structured JSON blackboard narrows the cost gap but still trails in success. These results suggest that structured shared state is helpful, and that transactional validation and write contracts provide additional gains beyond shared memory.

\subsection{Ablation Study}
\label{sec:ablations}

\begin{figure*}[t]
\centering
\includegraphics[width=0.95\linewidth]{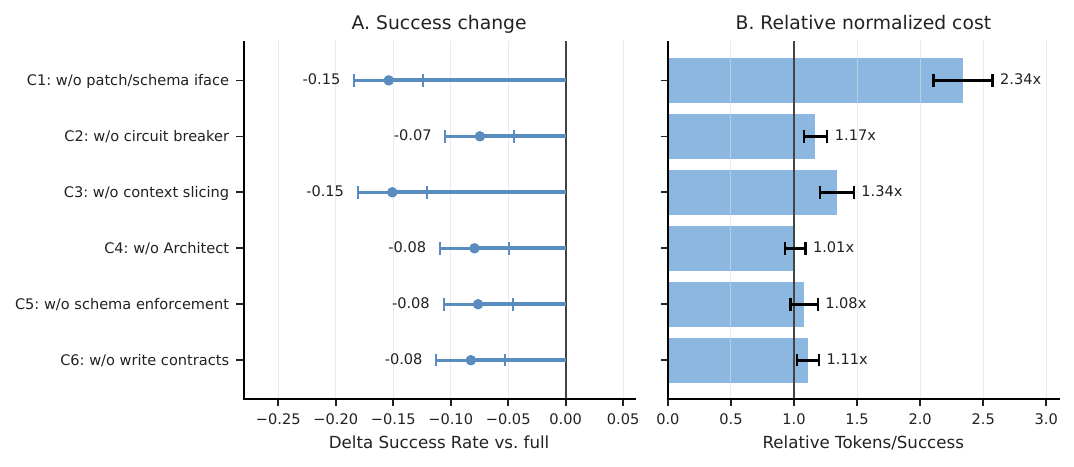}
\caption{Ablation impact relative to full \method{}. The left panel reports success-rate change relative to the full system. The right panel reports relative tokens per success, with 1.0 marking full \method{}.}
\label{fig:ablation-impact}
\end{figure*}

The strongest ablation effects come from removing the patch/schema interface and bounded context views. Figure~\ref{fig:ablation-impact} reports the change in success relative to full \method{} and the relative tokens per success.

C1 and C3 produce the largest drops, while other components show smaller but consistent effects. Removing the patch/schema interface causes the clearest cost failure, more than doubling tokens per success. Removing context slicing produces a comparable success drop, which supports the view-construction mechanism. The remaining ablations move in the expected direction, but their effects are smaller on this evaluation.

\subsection{Sensitivity Analyses}
\label{sec:sensitivity}

\begin{figure}[H]
\centering
\includegraphics[width=0.95\linewidth]{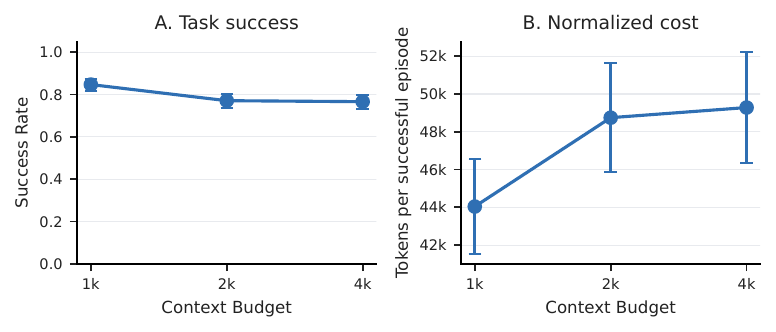}
\caption{Context budget sensitivity on ALFWorld.}
\label{fig:context-budget}
\end{figure}

\begin{figure}[htpb]
\centering
\includegraphics[width=0.95\linewidth]{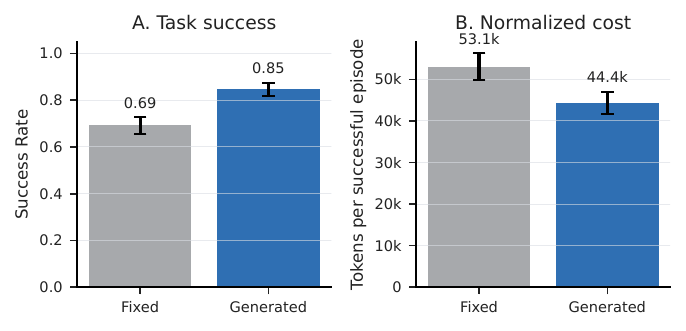}
\caption{Schema source sensitivity on ALFWorld.}
\label{fig:schema-source}
\end{figure}

We further examine two sensitivity factors: context budget measured in characters and schema source. Figure~\ref{fig:context-budget} reports the context-budget sensitivity, and Figure~\ref{fig:schema-source} reports the schema-source sensitivity.

The smallest tested context budget achieves the best success-cost profile. Increasing the budget does not improve success and leads to higher normalized cost. This result supports the bounded-view design: exposing more state to workers can introduce irrelevant context without improving local decision quality.

Generated task-specific schemas outperform fixed schemas in both success and normalized cost. This result supports using the Architect to construct task-specific blueprints in the current setting, while also showing that schema construction quality affects the reliability of the overall system.

\subsection{Fault Isolation and Termination}
\label{sec:fault-isolation}

Fault injection evaluates how each system handles malformed, unauthorized, semantic, and cyclic updates. For Invalid JSON, Bad Path/Type, Unauthorized Write, and False Claim, lower rates indicate less contamination or unsupported accepted content. For Cycle Halt, higher rates indicate more successful termination of repeated no-op or oscillatory trajectories.

\begin{table}[htpb]
\centering
\small
\resizebox{\linewidth}{!}{
\begin{tabular}{lccc}
\toprule
\textbf{Fault Type} & \textbf{Plain Blackboard} & \textbf{Structured JSON Blackboard} & \textbf{\method{}} \\
\midrule
Invalid JSON & 0.81 & 0.23 & 0.00 \\
Bad Path/Type & 0.76 & 0.18 & 0.00 \\
Unauthorized & 0.68 & 0.52 & 0.00 \\
False Claim & 0.70 & 0.66 & 0.43 \\
Cycle Halt & 0.12 & 0.39 & 0.96 \\
\bottomrule
\end{tabular}
}
\caption{Fault injection results over 200 injections per fault type.}
\label{tab:fault-injection}
\end{table}

\method{} has zero observed contamination for invalid JSON, bad paths/types, and unauthorized writes. False claims remain possible because a false claim can satisfy the schema. The high cycle-halt rate shows that the deterministic kernel can stop most repeated no-op or oscillatory trajectories.

\section{Conclusion}

We presented \method{}, a schema-grounded architecture for reliable and auditable LLM multi-agent collaboration. \method{} replaces open-ended inter-agent dialogue with validated JSON Patch mutations over a shared structured state, where an Architect defines the task schema, workflow rules, worker contracts, and context budgets, and a deterministic kernel validates proposed mutations, commits accepted updates transactionally, and records replayable logs. Across 630 matched ALFWorld episodes, \method{} achieves the strongest success-cost profile among the compared systems, solving 84.6\% of tasks and reducing tokens per successful task to 45.5k. Blackboard controls, ablations, and sensitivity analyses indicate that the gains come from the patch/schema interface, bounded context views, and transactional validation beyond shared memory alone. The diagnostic HotpotQA study further clarifies that schema-grounded mutation improves structural validity and auditability, while factual correctness still depends on evidence selection, verifier design, and task-specific semantic checks.

\section*{Limitations}
\label{sec:limitations}

\method{} provides structural reliability, so it does not guarantee semantic correctness. The deterministic kernel can reject malformed patches, unauthorized writes, and schema-violating state transitions, yet a schema-valid claim may still be false, incomplete, or unsupported. This limitation appears in the HotpotQA diagnostic, where structurally valid workflows can still produce unsupported claims. Factual tasks therefore require stronger evidence retrieval, verifier design, and human review for high-impact decisions.

The system also depends on blueprint quality, model behavior, and task setting. An overly sparse schema may block useful progress, while an overly permissive schema weakens role isolation. Our strongest evidence comes from ALFWorld, where task state and success conditions are clearly defined; more open-ended settings such as software engineering, scientific discovery, or long-form generation may require richer schemas and more complex validation. \method{} also introduces engineering overhead through schema construction, context slicing, patch validation, and transaction logging, making it most suitable when auditability, attribution, and state integrity are central requirements.

\section*{Ethical considerations}

Beyond the technical limitations above, auditable state transitions can make multi-agent systems easier to inspect, but they may also create false confidence. Users must not treat schema validity as factual correctness. Applications that affect people require evidence, provenance, and human review requirements for high-impact state transitions. Because transaction logs can contain sensitive intermediate information, deployments require access control, retention policies, and redaction mechanisms.

Our experiments use public research benchmarks, tools, and model APIs only for evaluation: ALFWorld, HotpotQA, LangGraph, Flock, and the Alibaba Cloud Bailian API. We cite the corresponding creators or providers and use these artifacts according to their public documentation, licenses, or access terms. We do not collect new human-subject data.

\bibliography{custom}

\newpage
\appendix

\section{\method{} Kernel Pseudocode}
\label{app:pseudocode}

Algorithm~\ref{alg:patchboard-kernel} summarizes the runtime loop. Worker calls can vary with the underlying model, but conditional on an accepted blueprint and proposed worker patches, validation, commit, logging, scheduling, and circuit decisions are deterministic.

The pseudocode makes explicit where the reliability boundary sits. A worker returns only a candidate patch, and the committed state changes only after \textsc{ValidPatch} checks syntax, authorization, patch application, post-state schema validity, and registered invariants. Rejected patches are still logged, preserving an audit trail without allowing invalid outputs to enter the shared state.

\begin{algorithm}[htpb]
\small
\caption{\method{} runtime kernel}
\label{alg:patchboard-kernel}
\begin{algorithmic}[1]
\Require user request $x$, blueprint meta-schema $\metaschema$
\Ensure committed state $\state$, transaction log $\mathcal{L}$
\State $b \gets \Call{Architect}{x}$
\If{\textbf{not} \Call{ValidBlueprint}{$b,\metaschema$}}
    \State \Return \Call{RejectBlueprint}{$b$}
\EndIf
\State $(\schema,\rules,\mathcal{C}) \gets \Call{Unpack}{b}$
\State $\state \gets \Call{InitialState}{x,\schema}$
\State $\mathcal{L} \gets [\,]$; $Q \gets \Call{InitialQueue}{\rules,\state}$
\While{$Q \neq \emptyset$ \textbf{and not} \Call{BudgetExceeded}{}}
    \State $(a,e) \gets \Call{Pop}{Q}$
    \State $\view^a \gets \Call{Slice}{\state,a,\mathcal{C}_a}$
    \State $\patch^a \gets \Call{Worker}{a,\view^a,e}$
    \State $(ok,\state',r) \gets \Call{ValidPatch}{\state,\patch^a,a,\schema}$
    \If{$ok$}
        \State $\state \gets \state'$
        \State $\mathcal{L} \gets \mathcal{L} \circ \Call{Commit}{a,e,\view^a,\patch^a,\state}$
        \State $E \gets \Call{Events}{\patch^a}$
        \State $Q \gets Q \circ \Call{Schedule}{E,\rules,\state}$
    \Else
        \State $\mathcal{L} \gets \mathcal{L} \circ \Call{RejectPatch}{a,e,\view^a,\patch^a,r}$
    \EndIf
    \State $u \gets \Call{CircuitPolicy}{\mathcal{L},\state,Q}$
    \State $(Q,\state) \gets \Call{ApplyPolicy}{u,Q,\state}$
\EndWhile
\State \Return $(\state,\mathcal{L})$
\end{algorithmic}
\end{algorithm}

Table~\ref{tab:kernel-ops} defines the deterministic operations used in Algorithm~\ref{alg:patchboard-kernel}. These operations mediate between model-generated patch proposals and committed shared state.

\begin{table}[htpb]
\centering
\small
\resizebox{\linewidth}{!}{
\begin{tabular}{lp{0.68\linewidth}}
\toprule
\textbf{Operation} & \textbf{Definition} \\
\midrule
\textsc{ValidPatch} & Checks JSON Patch syntax, allowed operation set, path existence, worker write contracts, stale-view \texttt{test} preconditions, post-application JSON Schema validity, and registered state-transition invariants. \\
\textsc{Slice} & Materializes the active task node, authorized read paths, relevant schema fragments, recent rejection reasons, summaries for large collections, and handles for expandable objects under a context budget $B$. \\
\textsc{CircuitPolicy} & Checks thresholds for consecutive invalid patches, repeated no-op proposals, repeated state hashes, and short state cycles, then returns a deterministic retry, repair, switch-worker, tighten-budget, or halt action. \\
\bottomrule
\end{tabular}
}
\caption{Deterministic kernel operations used in Algorithm~\ref{alg:patchboard-kernel}.}
\label{tab:kernel-ops}
\end{table}

\begin{figure*}[htpb!]
\centering
\includegraphics[width=0.9\linewidth]{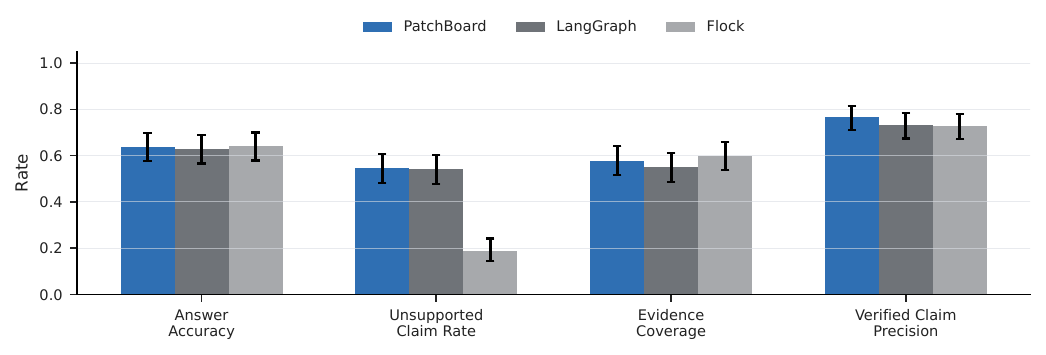}
\caption{Diagnostic HotpotQA results. All systems have similar answer accuracy, while unsupported-claim rates differ.}
\label{fig:hotpotqa-diagnostic}
\end{figure*}

The scheduler operates only on events emitted by committed patches. Downstream worker calls are therefore driven by accepted state transitions and workflow rules. The circuit policy is evaluated after each proposal so that repeated invalid edits, no-op loops, or short state cycles can be handled deterministically.

\section{Implementation and Reproducibility Details}

\subsection{Baseline System Settings}
\label{app:baseline_setting}

The LangGraph baseline uses the pure LangGraph runner in a supervisor--subagent configuration. Each environment turn is represented as a LangGraph state update. A supervisor node receives the current ALFWorld turn context and coordinates three tool-exposed subagents: a planner subagent, an action subagent, and a critic subagent. The planner produces or revises a local plan, the action subagent selects an admissible environment action, and the critic checks the candidate action before the supervisor finalizes the turn. This gives LangGraph a structured subagent workflow while leaving communication as tool-mediated message passing rather than schema-validated patch transactions.

The Flock baseline is run through a turn-level bridge that maps each ALFWorld state into a Flock orchestration call. This is a blackboard-family baseline in the sense that Flock agents communicate through typed artifacts in an orchestration context. In the reported comparison, the bridge uses a planner-and-executor workflow: the planner publishes a plan artifact from the local task state, and the executor consumes that artifact with the current turn context to return one admissible environment action. Flock receives the same observations, admissible-action interface, model, decoding configuration, step budget, and timeout as the other systems, but it does not use \method{}'s deterministic patch validator, role-specific write contracts, or transaction log.

\subsection{Run Configuration}
\label{app:exp-params}

Table~\ref{tab:exp-params} summarizes the run configuration used in the reported results, including the model backend, runtime hardware, benchmark scale, execution limits, and estimation settings. The main text describes the evaluated systems, controls, and ablation variants; this appendix records the concrete settings used for reproducibility.

\begin{table}[htpb]
\centering
\small
\resizebox{\linewidth}{!}{
\begin{tabular}{lr}
\toprule
\textbf{Setting} & \textbf{Value} \\
\midrule
Model provider & Alibaba Cloud Bailian API \\
Model version & qwen-plus-2025-12-01 \\
GPU & NVIDIA RTX 4090 \\
CPU & Intel Core i7-13700K \\
Memory & 32 GB RAM \\
Decoding temperature & 0 \\
Per-episode timeout & 60 seconds \\
\midrule
ALFWorld matched gamefiles & 126 \\
ALFWorld task types & 6 \\
Gamefiles per ALFWorld task type & 21 \\
Execution seeds per gamefile & 5 \\
ALFWorld episodes per system & 630 \\
Maximum environment steps per episode & 20 \\
\midrule
Context-budget settings & 1k, 2k, 4k characters \\
Circuit-threshold settings & 2, 4 \\
Loop-detection window & 3 state hashes \\
HotpotQA prepared validation examples & 240 \\
Fault injections per fault type & 200 \\
Bootstrap resamples & 10,000 \\
\bottomrule
\end{tabular}
}
\caption{Run configuration used for the reported results.}
\label{tab:exp-params}
\end{table}
\section{Diagnostic HotpotQA Results}
\label{app:hotpotqa}

The HotpotQA diagnostic is included to mark a limitation, not to claim transfer improvement. The three systems have similar answer accuracy, and Flock has the lowest unsupported-claim rate. This result supports the paper's boundary claim: schema-valid state transitions make coordination more auditable, while factual support still depends on evidence selection and verifier quality.

Figure~\ref{fig:hotpotqa-diagnostic} groups the semantic diagnostic metrics that are most relevant to this boundary. Answer accuracy shows whether the final answer is correct, while unsupported-claim rate measures whether the system introduced claims that were not backed by the prepared evidence. Evidence coverage and verified precision indicate how well the evidence-tracking fields support factual checking. The figure is therefore read as a semantic-support diagnostic rather than as an end-task win for \method{}.

\section{Running Example and Token Cost Breakdown}
\label{app:cost}

\begin{figure}[t]
\centering
\includegraphics[width=\linewidth]{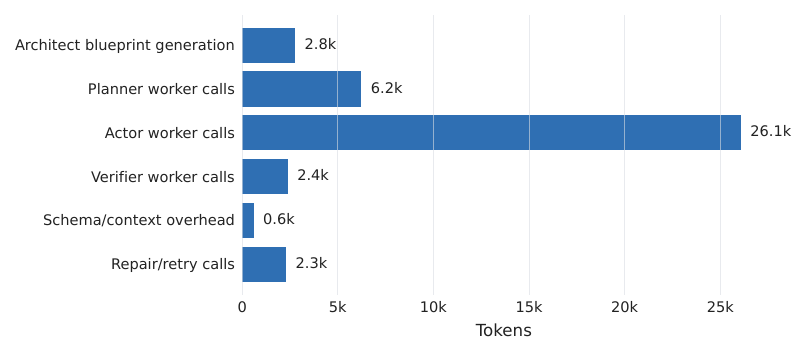}
\caption{Component-level token cost breakdown for a representative \method{} trajectory.}
\label{fig:cost-breakdown}
\end{figure}

Figure~\ref{fig:cost-breakdown} decomposes token usage for a representative \method{} trajectory on an ALFWorld clean-and-place task. The total accounted token usage is 40.3k. Actor worker calls are the largest component, accounting for 26.1k tokens, or 64.6\% of the total. This is expected because action selection is invoked repeatedly across environment turns and must condition on the current observation, admissible actions, recent state, and task objective.

The remaining costs span setup, planning, verification, and repair. Architect blueprint generation uses 2.8k tokens for one-time schema, contracts, workflow rules, and initial state construction. Planner calls consume 6.2k tokens to generate and revise local plans as observations change. Verifier calls use 2.4k tokens to check action admissibility and progress, while repair/retry calls use 2.3k tokens to handle inadmissible placement attempts.

Schema and context overhead accounts for only 0.6k tokens, or 1.5\% of the total. This category includes schema fragments, bounded state views, patch-format instructions, and state handles passed to workers. The breakdown supports an implementation-level interpretation: in this representative workflow, the main cost center is repeated action generation, while the schema and context machinery adds comparatively little token overhead. Future efficiency improvements should primarily target the number, size, or timing of actor calls, while preserving the validation and auditability benefits of schema-grounded state mutation.

\begin{figure*}[p]
\centering
\includegraphics[width=0.75\textwidth]{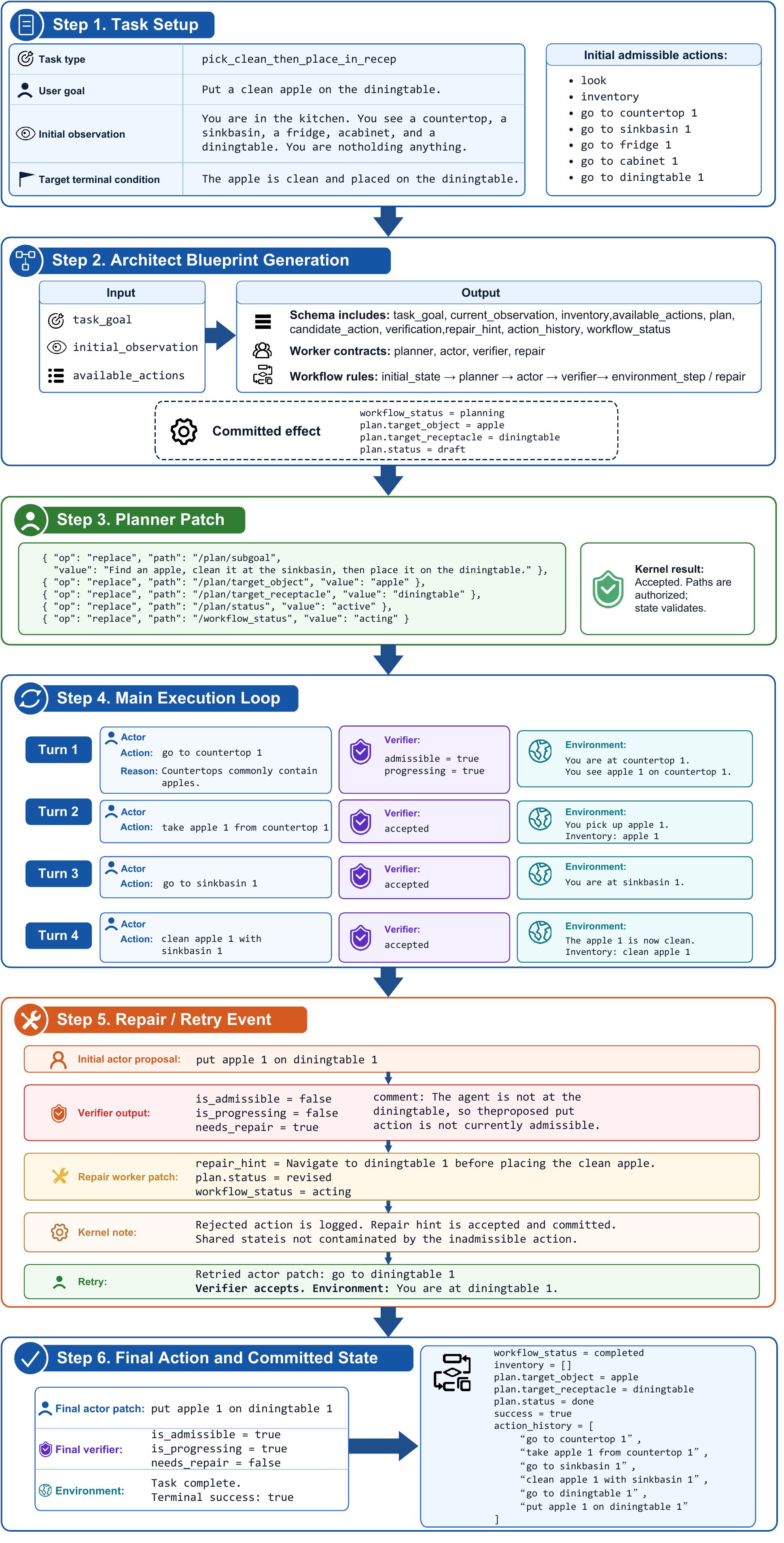}
\caption{Running example of \method{} on the ALFWorld clean-and-place task analyzed in Figure~\ref{fig:cost-breakdown}.}
\label{fig:running-example}
\end{figure*}

Figure~\ref{fig:running-example} gives the concrete execution trace behind the representative trajectory analyzed above. The task goal is to put a clean apple on the dining table. The task begins with the initial observation, admissible actions, and target terminal condition. The Architect then constructs a task blueprint containing the state schema, worker contracts, workflow rules, and context budgets. This blueprint initializes the shared state and defines which workers may read or modify each state region.

The planner first proposes a patch that fills the task subgoal, target object, target receptacle, and workflow status. The kernel accepts the patch only after checking that the edited paths are authorized and that the resulting state satisfies the task schema. During the main execution loop, the actor proposes environment actions, the verifier checks admissibility and task progress, and accepted actions are executed in the ALFWorld environment. The trace illustrates how environment feedback is converted into committed state updates, allowing later worker calls to condition on the validated state trajectory rather than on an unstructured dialogue history.

The repair step highlights the validation boundary enforced by \method{}. After cleaning the apple, the actor initially proposes placing it on the dining table before the agent has navigated there. The verifier marks this candidate action as inadmissible and requests repair. The rejected action is logged, while the repair hint is accepted as a separate state update. The subsequent retry navigates to the dining table, after which the final placement action completes the task. This example shows how invalid intermediate proposals can remain auditable without contaminating the committed shared state.
\end{document}